\documentclass[runningheads]{llncs}
\usepackage{graphicx}
\usepackage{epsfig}
\usepackage{comment}
\usepackage{amsmath,amssymb} 
\usepackage{nicefrac}
\usepackage{tabulary,multirow,overpic,xcolor}
\usepackage{color}
\usepackage{subfiles}
\usepackage{mmstyles}
\usepackage[ruled,vlined]{algorithm2e}
\usepackage[width=122mm,left=12mm,paperwidth=146mm,height=193mm,top=12mm,paperheight=217mm]{geometry}

\usepackage[caption=false]{subfig}
\usepackage{epstopdf}
\def\x{$\times$}
\definecolor{demphcolor}{RGB}{144,144,144}
\newcommand{\demph}[1]{\textcolor{demphcolor}{#1}}
\newlength\savewidth\newcommand\shline{\noalign{\global\savewidth\arrayrulewidth
		\global\arrayrulewidth 1pt}\hline\noalign{\global\arrayrulewidth\savewidth}}

\usepackage[pagebackref=true,breaklinks=true,letterpaper=true,colorlinks,bookmarks=false]{hyperref}

\begin{document}
\pagestyle{headings}
\mainmatter
\def\ECCVSubNumber{1701}  

\title{PV-NAS: Practical Neural Architecture Search for Video Recognition} 

\titlerunning{Abbreviated paper title}
%
\author{Zihao Wang\inst{1}\thanks{The work was done during his internship at SenseTime Research} \and
Chen Lin\inst{1,2} \and
Lu Sheng\inst{3} \and
Junjie Yan\inst{1} \and 
Jing Shao\inst{1}\thanks{Corresponding author}}

\authorrunning{F. Author et al.}
%
\institute{SenseTime Research \and
Engineering Science, University of Oxford \and
College of Software, Beihang University \\
\email{zihaowang.cv@gmail.com} \  \  \  \   \email{chen.lin@eng.ox.ac.uk} \\
 \email{lsheng@buaa.edu.cn} \  \  \   \email{\{yanjunjie,shaojing\}@sensetime.com}\\}
\maketitle

\begin{abstract}

Recently, deep learning has been utilized to solve video recognition problem due to its prominent representation ability.
Deep neural networks for video tasks is highly customized and the design of such networks requires domain experts and costly trial and error tests.
Recent advance in network architecture search has boosted the image recognition performance in a large margin.
However, automatic designing of video recognition network is less explored.
In this study, we propose a practical solution, namely Practical Video Neural Architecture Search (PV-NAS).
Our PV-NAS can efficiently search across tremendous large scale of architectures in a novel spatial-temporal network search space using the gradient based search methods. 
To avoid sticking into sub-optimal solutions, we propose a novel learning rate scheduler to encourage sufficient network diversity of the searched models.
%
Extensive empirical evaluations show that the proposed PV-NAS achieves state-of-the-art performance with much fewer computational resources.
%
%
1) Within light-weight models, our PV-NAS-L achieves 78.7\% and 62.5\% Top-1 accuracy on Kinetics-400 and Something-Something V2, which are better than previous state-of-the-art methods (\ie, TSM) with a large margin (4.6\% and 3.4\% on each dataset, respectively), and 2) among median-weight models, our PV-NAS-M achieves the best performance (also a new record) in the Something-Something V2 dataset.

\vspace{-10pt}
\keywords{Video Recognition, Neural Architecture Search, Deep Learning}
\vspace{-10pt}
\end{abstract}
\vspace{-10pt}
\section{Introduction}

Deep convolutional neural network (CNN) has achieved remarkable performances in the task of video recognition~\cite{carreira2017quo,tran2018closer,qiu2017learning,wang2016temporal,zolfaghari2018eco,zhou2018temporal}.
However, most of the previous works directly borrow the deep learning models designed for 2D image classification with minor modifications.
For example, the well-known I3D~\cite{carreira2017quo} model replaces several 2D convolution operations with 3D convolutions and at the meanwhile remains the inception-block structure as that in Inception-V1~\cite{inception}.
Therefore, it is promising if we can find a network architecture that conforms the nature of video representation, which could potentially increase the upper bound of video recognition task.
However, it is difficult to use standard trial and error pipeline to craft a customized network for video recognition due to its huge computation cost.

\begin{figure}[t]
\begin{center}
   \includegraphics[width=0.6\linewidth]{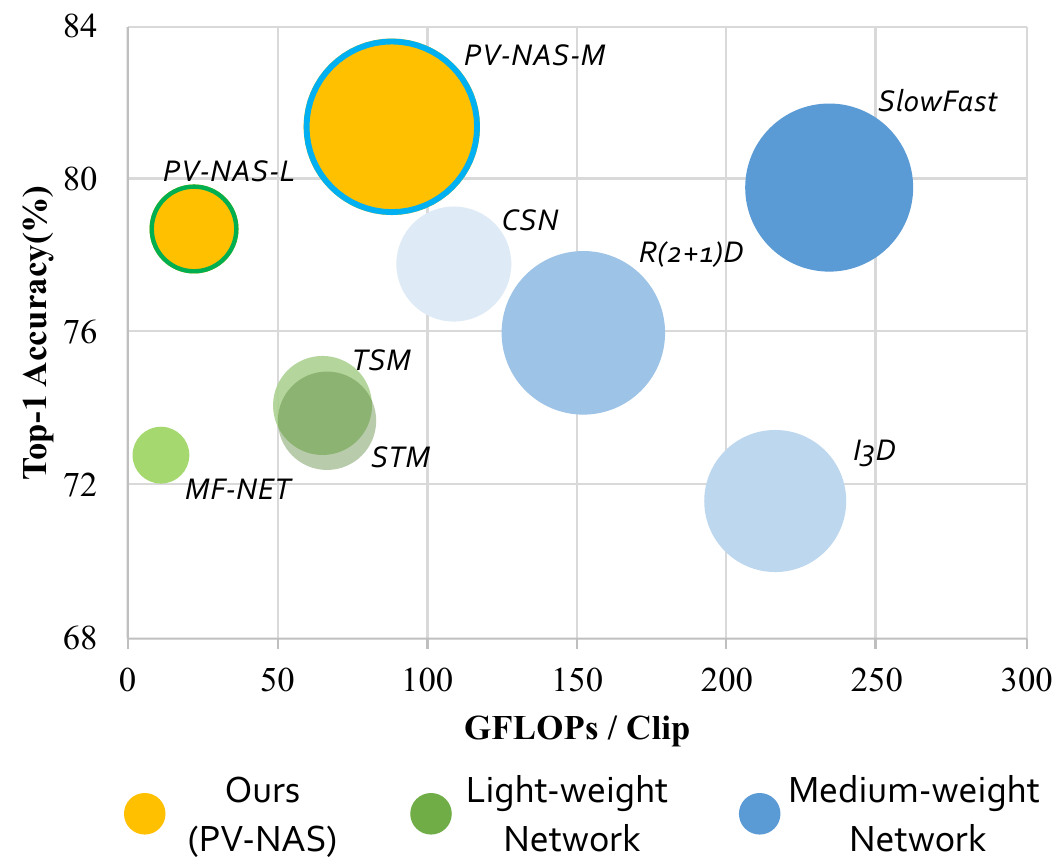}
\end{center}
\vspace{0pt}
   \caption{The proposed PV-NAS (marked in warm color) outperforms and has better accuracy-cost trade-off than all previous approaches for video recognition on Kinetics-400. The light-weight and medium-weight networks are represented by green and blue, respectively. The size of circles indicate the number of parameters.}
\label{fig:resultplot}
\vspace{0pt}
\end{figure}

Interestingly, recent advances in efficient neural architecture search (NAS) provide a feasible way for this problem.
Automatically searched network architectures have achieved competitive performance on 2D tasks such as image classification~\cite{liu2018darts,wu2019fbnet,pham2018efficientnas,cai2018proxylessnas,liu2018progressive,chen2018reinforced} and object detection \cite{nasfcos,liu2019cbnet,chen2019detnas,peng2019efficient}.
And there have been a few pioneer works~\cite{Piergiovanni_2019_ICCV,assemble_2019} that attempt to apply NAS in video recognition.
However their search cost are expensive, which requires thousands of GPU days to evaluate the mutated candidates.
The practicality of these methods tends to be marginal considering their huge computational budget.

\begin{figure*}[t]
\begin{center}
   \includegraphics[width=0.7\linewidth]{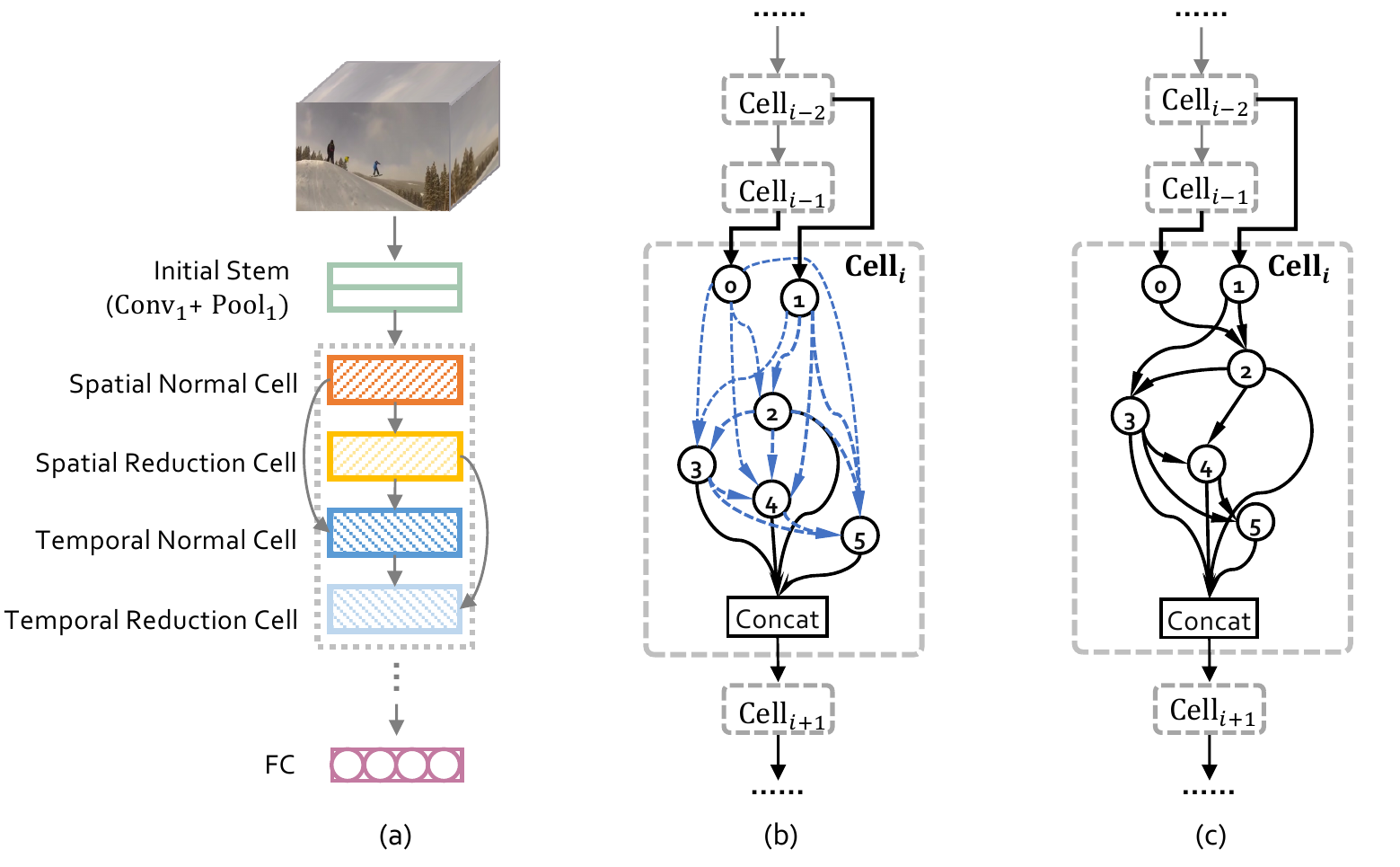}
\end{center}
\vspace{-10pt}
      \caption{An overview of PV-NAS. (a) The overall neural architecture with four cell categories. We optimize the shared architecture parameters for each cell to learn an optimal architecture. (b) The continuous relaxation of a cell. The cell is initialized as a dense DAG and each blue-marked edge is a mixed operation.(c) An instance of the searched temporal normal cell architecture. Withe the optimization, we remain two strongest operations for each intermediate node.
      \setlength{\belowcaptionskip}{-10pt}
      }
\label{fig:intro_compare}
\vspace{0pt}
\end{figure*}

In this study, we establish a practical pipeline to search for customized networks for video recognition with the differential NAS techniques, namely the Practical Video NAS (PV-NAS).
%
We provide a reference design of the search space which has been thoroughly tested and refined.
Specifically, our search space contains multiple cells, \emph{Spatial Normal Cell} (S-NC), \emph{Spatial Reduction Cell} (S-RC), \emph{Temporal Normal Cell} (T-NC), and \emph{Temporal Reduction Cell} (T-RC).
The spatial cells are intended to extract 2D image information, which only contain 2D operations.
The temporal cells expand the search space with additional temporal operations, which aims at capturing temporal motion features.
%
This advantage of this type of spatio-temporal cells is that by manipulating the stacking method of spatial cells and temporal cells, we can constraint the possible solution space to contain more performant and efficient models.
To achieve reduce the search cost, we borrow the idea of differential network architecture optimization from NAS literature ~\cite{liu2018darts,cai2018proxylessnas,wu2019fbnet}, where we conduct a continuous relaxation to our proposed search space and optimize the architecture with gradient descent.
However, we observe that a na\"ive combination of differential NAS and video recognition would leads to mediocre results.
We locate this false case and relate it to the optimization dynamics of architecture.
A customized learning rate scheduler is showed helpful, by encouraging the discovery of diverse model architectures.
%
We set up a reference implementation on NAS for video tasks which is computation friendly and empirically effective. Our searched models surpass the recent state-of-the-art hand-crafted networks by a considerable margin.
%
For light-weight network comparison, we show that PV-NAS-L with 19.65M parameters (or 22.14 GFlops) is able to achieve 78.7\% Top-1 accuracy on Kinetics~\cite{kay2017kinetics}, which achieves 4.6\% performance gain compared to the state-of-the-art result obtained by the hand-crafted network TSM~\cite{lin2019tsm} with a similar scale of network parameters.
It also achieves 62.5\% Top-1 accuracy when transferred to Something-Something action recognition dataset~\cite{goyal2017something}, which also outperforms TSM with 3.4\%.
For medium-weight setting, we obtain a new state-of-the-art architecture PV-NAS-M with 11 layers and 77.15M parameters.
As shown in Fig.~\ref{fig:intro_compare}, on the Kinetics-400 dataset, our PV-NAS-M achieves 81.4\% Top-1 accuracy, which surpass the previous state-of-the-art SlowFast~\cite{Feichtenhofer_2019_ICCV} 

We aimed to draw attentions to the practicality of the application of NAS methods, especially for video recognition tasks. Our contributions are listed as follows:

1) \textit{Practical Search Space} - We provide a reference design of the search space for video tasks which has been thoroughly tested and refined. 

2) \textit{Solution to the False Case} - We identify the false case of naive implementation of differential based video NAS and provide a empirically effective solution to it.

3) \textit{Reference Models} - The searched models PV-NAS-L and PV-NAS-M surpass the performance records on Kinetics-400 and Something-Something V2 dataset under adverse conditions(less frames for training \& testing). Notably, our models contains fewer parameters and consumes less computation. We believe this would provide new clues for future video network architecture design.

\section{Related Work}

\subsection{Video Recognition}

In the past few years, video recognition has seen a major paradigm shift, which involved moving from 2D-oriented CNN \cite{simonyan2014two,Karpathy_2014_CVPR,lin2019tsm} to 3D-based CNN \cite{tran2015learning,carreira2017quo,Xie_2018_ECCV,Qiu_2017_ICCV,assemble_2019,Feichtenhofer_2019_ICCV}.

A typical 2D-based CNN for video recognition often takes two or several separate branches for learning spatial- and temporal-features respectively, such as Two-stream CNN \cite{simonyan2014two} and Slow-Fusion CNN \cite{Karpathy_2014_CVPR}. Another work, TSN \cite{simonyan2014two}, is proposed to extract averaged features from strided sampling frames and fused their features at multiple stages to capture long-term spatio-temporal features. Since the extractions of appearance and dynamic representations have no interactions, the performances are not satisfied.
Besides, these variants, simply transfer 2D operations for image-related tasks to video recognition, are of low cost in memory and calculation, but they inevitably sacrifice the descriptive ability for the inherent temporal patterns.

Alternatively, more experts adopt 3D-CNN for video recognition by extending appearance feature learning in a 2D CNN to its counterpart to simultaneously learn appearance and motion features on the input 3D video volume~\cite{tran2015learning,carreira2017quo}. Tran \etal~\cite{tran2015learning} proposed a 3D-CNN based on the VGG model, named C3D, to learn spatio-temporal features. While Carreira and Zisserman~\cite{carreira2017quo} proposed to inflate all the 2D convolution filters in an Inception V1 model into 3D convolutions.
However, substituting all 2D kernels with 3D kernels make the model complexity increases dramatically.

Recent works, such as TSM~\cite{lin2019tsm}, tried to enjoy the same computation as 2D CNNs and meanwhile enable the same spatio-temporal modeling ability.
Similar works decomposed 3D convolution operations into a Pseudo-3D convolutional block as in P3D~\cite{Qiu_2017_ICCV} or factorized convolutions as in R(2+1)D~\cite{tran2018closer} or S3D~\cite{Xie_2018_ECCV}.
However, these works need experienced researchers and always have bad generalization. Thus constructing an efficient but also effective networks with both 2D and 3D operations still remains an open problem.

The proposed PV-NAS model overcomes the aforementioned limitations. With innovative spatial and temporal search space, the automated designed network is capable of achieving the state-of-the-art results and at the meantime taking less computation cost than other NAS-related method.

\subsection{Neural Architecture Search}

Recently, automatic machine learning techniques have achieved remarkable progresses, such as augmentation \cite{cubuk2018autoaugment,lin2019online}, loss function \cite{li2019lfs} and neural architecture search(NAS) \cite{zoph2016neural,brock2017smash,liu2018darts,guo2019single}.
Among which, NAS has been shown compatible or even outperformed to the hand-craft network architectures.
Early NAS works can be typically considered as an agent-based explore and exploit process. Two widely used agents are reinforcement learning agent \cite{zoph2016neural} and evolution agent \cite{real2017large}.
The generated networks are usually trained from scratch on the original or proxy datasets.
These methods are facing problem of expensive computational costs.
To overcome this issue, \cite{pham2018efficientnas} proposes to adapt a weight sharing strategy that reduces the computation.
\cite{liu2018darts} first imposes a continuous relaxation on discrete network search space with a real-valued gate for each operator and alternately train operator weights and parameters for gates by back-propagation.
\cite{cai2018proxylessnas} follows the differential optimization pipeline, while revisits the definition of architecture parameters in a probabilistic point of view.
Several efficient NAS methods disentangle the training of share parameters and the searching for high performance models into two stages \cite{bender2018understanding,guo2019single}.


\section{Practical Video Neural Architecture Search  (PV-NAS)}

In this section, we first introduce the construction of the proposed cell-based spatial-temporal network search space for video input in Sec.~\ref{sec:searchspace}.
%
%
In Sec.~\ref{sec:optimization}, we discuss how we apply continuous relaxation to the discrete search space. And how we jointly optimize the architecture parameter as well as network weights with gradient descent based on a novel learning scheduler in Sec.~\ref{secsec:slr}.

\subsection{Spatial-Temporal Search Space}\label{sec:searchspace}

Cell based architectures have been heavily investigated in NAS literature \cite{zoph2016neural,real2017large,liu2018darts}. Thus, we adopt the cell-based design paradigm in our search space.
Following the differential NAS approach~\cite{liu2018darts}, we employ a cell-based spatio-temporal search space in which the network is composed by stacking several categories of \textit{cell}s.
We first briefly introduce the definitions of cells. Then we give a detail description of our spatio-temporal search space for video tasks.
\subsubsection{Cell}
We define cell as a basic building block for our models. A cell could be represented as a direct acyclic graph, which contains $N$ (ordered) nodes $\mathcal{X}=\{x_0, x_1, ..., x_N\}$ and a set of edges $\mathcal{E}=\{{(i,j)\}}$, if node $i$ and node $j$ are connected.
Notably for CNN architecture, each node $x_i\in\mathbb{R}^{C{\times}T{\times}W{\times}H}$ represents a intermediate feature map and each edge $(i,j)$ takes a value from a predefined operation set $\mathcal{O}$ as the corresponding operation $o^{(i,j)}$. The example of the operation set refers to \ref{operationset1}. If $o^{(i,j)}=o_k$, we can conclude that it is the $k$-th operation transforms the $j$-th feature map to the $i$-th. If a node has multiple incoming edges, we would simply conduct a summation over all of them.
\begin{equation}
    x_i = \sum_{(i,j)\in \mathcal{E}}{o}^{(i,j)}(x_j).
\end{equation}
We define the output of a cell as the concatenation of all its intermediate nodes $\{x_2, x_3, ..., x_n\}$ along the channel dimension.
Furthermore, following previous works, we fix the the in-degree of each node to 2 except the first two nodes $x_0$ and $x_1$, which are the input nodes. The input nodes are set as the output of two preceding cells.

\subsubsection{Spatial Cell}
We define the spatial cell as the cell only contains spatial operators.
Similar as the cells used in NAS for image recognition, there are two categories of cells in Spatial Cell: the \textit{normal cell} (S-NC) and \textit{reduction cell} (S-RC). Both the two categories share a common operation set $\mathcal{O}_s$. However, the reduction cell set the stride as 2 for all the operations which transforms the feature map from an input node to produce an output with a reduced resolution.
\emph{Spatial Operation Set}: The corresponding operation set $\mathcal{O}_s$ is a subset of a common used operation set that only contains 2D operations: 

\begin{table}[h] \label{operationset1}
    \setlength{\tabcolsep}{0.1em} 
    \centering
    \begin{tabular}{l@{~~~~~~~~~~~~}l}
    $\bullet\;$ Identity  & $\bullet\;$ Zero \\
    $\bullet\;$ 1$\times$3$\times$3 ave pooling  &$\bullet\;$ 1$\times$3$\times$3 max pooling  \\ 
    $\bullet\;$ 1$\times$3$\times$3 separable conv &$\bullet\;$ 1$\times$3$\times$3 dilated separable conv  \\
    \end{tabular}
\end{table}
%

\subsubsection{Temporal Cell}
We define the temporal cell as the cell contains temporal operators as well.
Temporal information has been proved crucial for video recognition tasks.
However, directly adding temporal operations to the operation set of Spatial Cells would lead to bad performance. 
This is due to the fact that using temporal convolutions in earlier layers degrades the accuracy, first observed by Feichtenhofer~\etal~\cite{Feichtenhofer_2019_ICCV}. 
To eliminate these sub-optimal architectures in our search space, we propose two specified temporal cells (\ie, \textit{Temporal Normal Cell} (T-NC) and \textit{Temporal Reduction Cell} (T-RC)) for the purpose of processing the temporal information.
Thus, we can avoid  temporal convolutions in earlier layers by predefining the order of different cells.

\emph{Temporal Operation Set}\label{secsec:temporalspace}:
We introduce two kinds of temporal operations to expand the operation set of the spatial cell: 3$+$3$\times$3 and 3$+$1$\times$1 temporal convolutional layers~\cite{tran2018closer}.
The shared operation set for temporal cells $\mathcal{O}_t$ is:
\begin{table}[h] \label{operationset2}
    \setlength{\tabcolsep}{0.1em} 
    \centering
    \begin{tabular}{l@{~~~~~~~~~~~~}l}
    $\bullet\;$ Identity  & $\bullet\;$ Zero \\
    $\bullet\;$ 1$\times$3$\times$3 ave pooling  &$\bullet\;$ 1$\times$3$\times$3 max pooling  \\ 
    $\bullet\;$ 1$\times$3$\times$3 separable conv &$\bullet\;$ 1$\times$3$\times$3 dilated separable conv  \\
    $\bullet\;$ 3$+$3$\times$3 temporal conv  & $\bullet\;$ 3$+$1$\times$1 temporal conv
    \end{tabular}
\end{table}

\begin{figure}[t]
\begin{center}
   \includegraphics[width=0.5\linewidth]{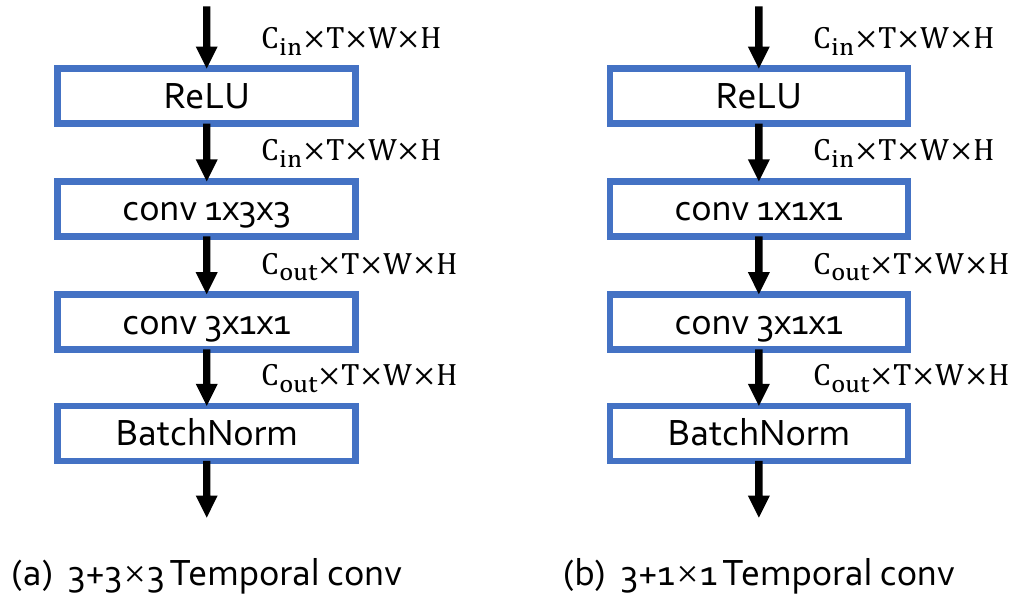}
\end{center}
   \caption{Temporal operation set. (a) The $3+3{\times}3$ temporal convolution, splits the computation of a dense 3D convolution into a 2D spatial convolution and a following 1D temporal convolution. (b) The $3+1{\times}1$ temporal convolution, utilizes $1\times1$ spatial convolution, which reduce the computation compared with the former operation.
   \setlength{\belowcaptionskip}{-10pt}
   }
\label{fig:temporal}
\vspace{0pt}
\end{figure}

As shown in Fig.~\ref{fig:temporal}, each temporal operation, \ie, 3$+$1$\times$1 or 3$+$3$\times$3 temporal convolutional layer follows a pattern that a spatial 2D convolution pluses a 1D convolution on the temporal dimension (\ie,  \texttt{ReLU-Conv2D-Conv1D-BN}).
%

In our implementation, each cell has $2$ input nodes and $4$ intermediate nodes.
Thus, the whole search space includes $\prod_{k=1}^{4}{\frac{(k+1)k}{2}{\times}(5^2)}\approx7\times10^7$ possible architectures for spatial cells and $\prod_{k=1}^{4}{\frac{(k+1)k}{2}{\times}(7^2)}\approx10^9$ possible architectures for temporal cells.
Without considering graph isomorphism, we have around $(7{\times}10^{7})^2{\times}10^{9^2}\approx 4.9\times10^{33}$ final network architectures in our search space.

\subsection{Differentiable Architecture Optimization}
\label{sec:optimization}

\subsubsection{Continuous Relaxation}
\label{secsec:relax}
Since exhaustively searching all the possible architectures in the discrete search space is almost impossible, it is more practical to employ a super cell for each cell category to optimize the cell structure in a continuous space. As shown in Fig.\ref{fig:intro_compare}(b), a super cell is the union of all the possible edges with all the candidate operations of the corresponding cell category.
Following~\cite{liu2018darts}, a mixed operation $\bar{o}^{(i,j)}$ is used to represent the union of all the available operations for edge $(i,j)$. Also, for the candidate operations in that edge $o^{(i,j)}\in \bar{o}^{(i,j)}$, we use a set of architecture parameters $\{\alpha_o^{(i,j)}\}$ to compute the importance of each single operation.

Suppose a given super cell with its corresponding architecture parameter $\valpha$ and operation set $\mathcal{O}$.
The mixed operation could relax the categorical choice of operations to a summation over all the candidates $o\in \mathcal{O}$, which are weighted by a dedicate set of  architecture parameters ${\alpha_o^{(i,j)}}$ as,
\begin{equation}
\bar{o}^{(i,j)}(x_i) = \sum_{o\in\mathcal{O}}{\frac{\exp(\alpha_{o}^{(i,j)})}{\sum_{o'\in \mathcal{O}}{\exp(\alpha_{o'}^{(i,j)})}}o^{(i,j)}(x_i)}, ~\text{for edge}~(i,j)\in\cE
\end{equation}
$\valpha = \{ \alpha_o^{(i,j)} \}$ is the set of architecture parameters.
%
To solve an optimal set $\valpha$, we need to optimize the training loss simultaneously with respect to the architecture parameters and network weights $\vw = \{w(\bar{o}^{(i,j)})\}$.
The candidate network could be inferred from the solved architecture parameters.
Specifically, for each edge, we first remove all the candidate operations except the operation with a maximum score.
Then we keep two preceding edges for each nodes according to the operation scores in the remained edges.
The score of an operation is defined as ${\exp(\alpha_{o}^{(i,j)})}/{\sum_{o'\in \mathcal{O}}}{\exp(\alpha_{o'}^{(i,j)})}$.
%
\subsubsection{One-Level Approximation}\label{secsec:onelevel}
\begin{figure}[t]
\begin{center}
   \includegraphics[width=0.75\linewidth]{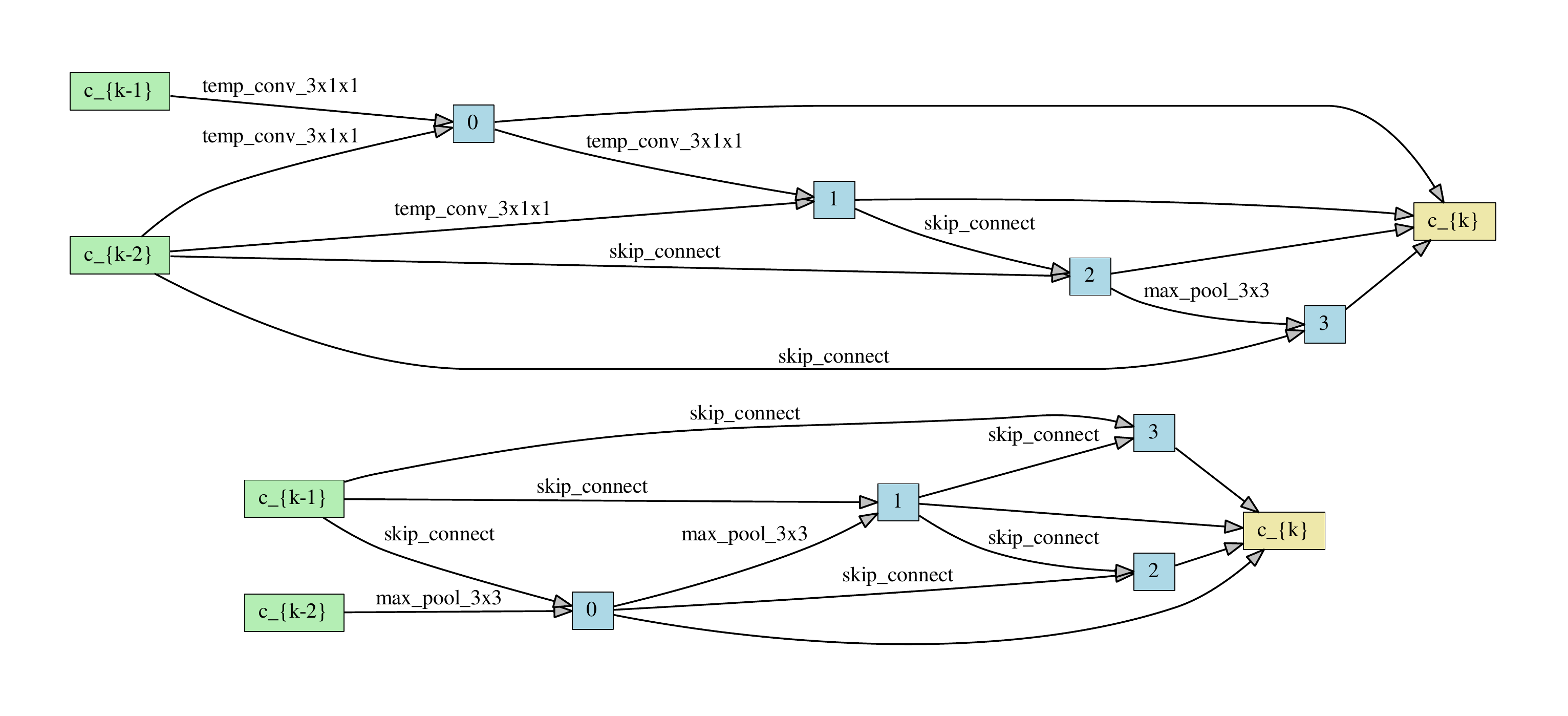}
\end{center}
   \caption{Those are two examples of searched temporal normal cells while using cosine learning rate scheduler. Most of the operations are learned to be parameter-free since that would be easier to optimize the loss function.}
\label{fig:cosarch}
\vspace{0pt}
\end{figure}
Most works on differential architecture search formulated NAS as a bi-level optimization problem.
The inner level solves for the optimal weights associated with the architecture parameters $\valpha$, defined as $\vw^*(\valpha) = \arg\min_w \enskip \cL_{train}(\vw, \valpha)$. The outer level solves for the architecture parameter $\alpha$ by minimizing the validation loss: 
\begin{align}
	\min_\alpha \quad & \mathcal{L}_{val}(\vw^*(\valpha), \valpha), \label{eq:outer} \\
	\text{subject to} \quad & \vw^*(\valpha) \arg\min_\vw \enskip \mathcal{L}_{train}(\vw, \valpha). \label{eq:inner}
\end{align}
Since it is hard to optimize this bi-level problem due to the inner optimization, previous methods usually employ a single step approximation to simplify the inner optimization, where
\begin{equation}
    \nabla_{\valpha} \mathcal{L}_{val}(\vw^*(\valpha), \valpha) \approx \nabla_{\valpha} \mathcal{L}_{val}(\vw - \xi \nabla_{\vw} \mathcal{L}_{train}(\vw, \valpha), \valpha) \label{eq:single}
\end{equation} 
However, such approximated problem is still not affordable for video recognition tasks.
Firstly, training video networks is usually computationally expensive with a high memory cost.
Since the inner and outer optimizations usually employ different optimizers, thus the overall optimization would double the memory cost and the computational cost, making the training process can not be afforded.
Secondly, the video networks are more sensitive to the distribution of datasets. Since we only sample a 8-frame clips for each video, the validation set should be carefully split from the training set to keep the number of videos of each class balanced.
To resolve aforementioned problem, we adopt one-level optimization strategy: directly optimize the network weights $w$ and the architecture parameter $\alpha$ with the training loss,
\begin{align}
	\min_{\valpha} \quad & \mathcal{L}_{train}(\vw^*(\valpha), \valpha), \\
	\text{where} \quad & \vw^*(\valpha) \approx \vw - \xi \nabla_{\vw} \mathcal{L}_{train}(\vw, \valpha). \label{eq:onelevel}
\end{align}
This is implemented by alternatively optimizing $\valpha$ and $\vw$ by gradient descent:
\begin{align}
\valpha_t & = \valpha_{t-1} - \delta_t  \partial_{\valpha} \cL_{train}(\vw_{t-1},\valpha_{t-1}) \\
   \vw_t & = \vw_{t-1} - \eta_t \partial_w \cL_{train}(\vw_{t-1},\valpha_{t}) 
\end{align}
where the $\delta_t$ and $\eta_t$ are the learning rates for $\valpha$ and $\vw$ respectively.

\subsection{Scheduled Learning Rate}\label{secsec:slr}

\begin{algorithm}[t]
\SetAlgoLined
    \caption{\footnotesize PV-NAS Algorithm}\label{alg:pvnas}
        Create a mixed  operation $\bar{o}^{(i,j)} (x)$ parameterized by  $\alpha_{o}^{(i,j)}$ for each edge $(i,j)$\;
        \While { $t <  T $}{ 
        1. Update architecture parameter $\valpha_t$ by descending $\partial_{\valpha} \cL_{train}(\vw_{t-1},\valpha_{t-1})$\;
        2. Update weights $\vw_t$ by descending $\partial_\vw \cL_{train}(\vw_{t-1},\valpha_{t})$\;
        \If{ $t \mod s \equiv 0$}{Decay the learning rate of weights by $d$\;}
        }
        Set optimal $\valpha^*=\valpha_T$\;
        Derive the final architecture with the learned $\valpha^*$\;
\end{algorithm}

We follow the previous works~\cite{liu2018darts,Chen_2019_ICCV} to use the cosine scheduler to anneal the learning rate from 0.1 to 0.0001.
However, we observed that architectures converge to bad local minima in which the searched architectures only contain abundant skip-connection operations or none operation.
Multiple runs conform that the tendency to fall into bad local minimum consistently emerges.
We show two example of the resulted bad case in Fig.~\ref{fig:cosarch}.
In order to escape the bad local minima, we investigate the effect of different learning rate scheduling.

Firstly, we tried constant learning rate 0.1.
Surprisingly, constant learning rate is robust to bad local minimum: the architecture parameters converge to these architectures do not contain too many abundant operations.
We explain the tendency to bad solution phenomenon as follows: the convergence speed of model parameters and architecture parameters is fundamentally different, a small learning rate would leads to fast converge of architecture parameters when the model parameters are still at the early stage of learning.
Terrible model parameters forces the architecture to converge to models that only perform good in their early stage of training (which is exactly the kind of architecture with less parameters).

Based on this intuition, we replace the cosine learning rate scheduler (for the architecture parameters) in previous methods by a step learning rate decay strategy, which does not decay the learning rate until the architecture parameters converges.
We found that the schedule that decays by $10$ times for sufficient epochs usually finds promising architectures. The whole procedure is illustrated as Alg.~\ref{alg:pvnas}.


	\newcommand{\blocks}[3]{\multirow{3}{*}{\(\left[\begin{array}{c}\text{1$\times$1$^\text{2}$, #2}\\[-.1em] \text{1$\times$3$^\text{2}$, #2}\\[-.1em] \text{1$\times$1$^\text{2}$, #1}\end{array}\right]\)$\times$#3}
	}
	\newcommand{\blockt}[3]{\multirow{3}{*}{\(\left[\begin{array}{c}\text{\underline{3$\times$1$^\text{2}$}, #2}\\[-.1em] \text{1$\times$3$^\text{2}$, #2}\\[-.1em] \text{1$\times$1$^\text{2}$, #1}\end{array}\right]\)$\times$#3}
	}
	\newcommand{\outsizes}[7]{\multirow{#7}{*}{\(\begin{array}{c} \text{\emph{Slow}}: \text{#1$\times$#2$^\text{2}$}\\[-.1em] \text{\emph{Fast}}: \text{#4$\times$#5$^\text{2}$}\end{array}\)}
	}

\begin{figure}[t]
\begin{center}
   \includegraphics[width=0.6\linewidth]{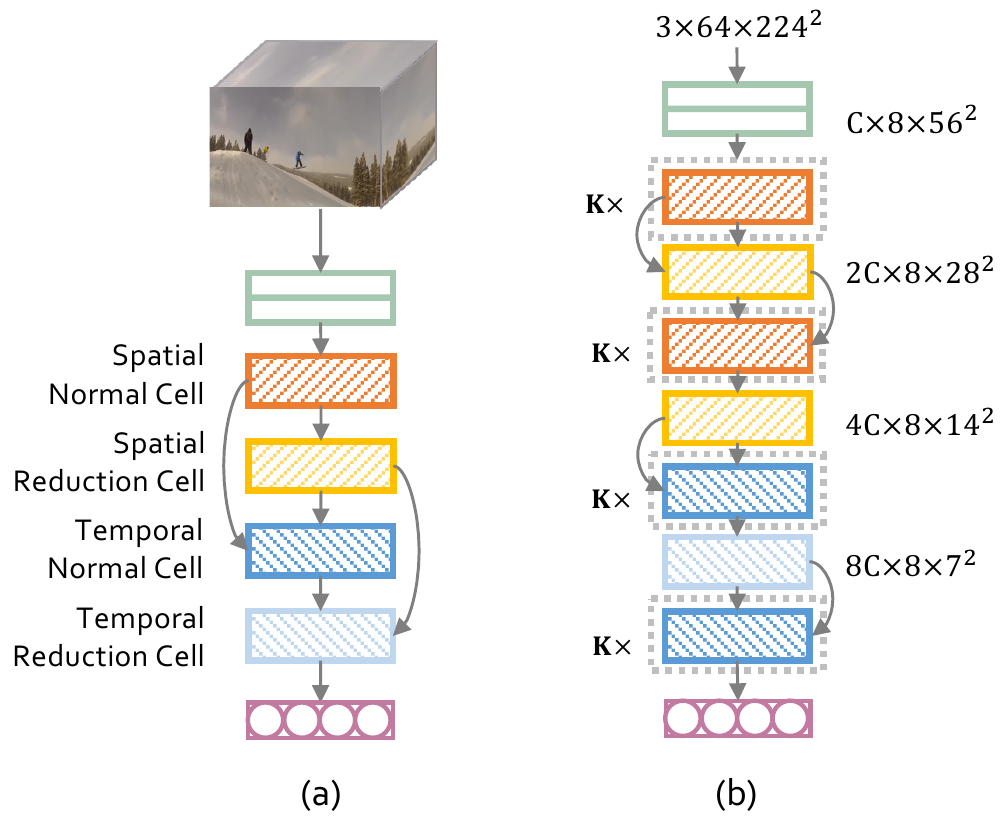}
\end{center}
      \caption{The stacked cell architecture used for search and evaluation. (a) For the search stage, we stack each category of cell only once to reduce the computation burdens and memory expense (b) For the retrain stage, we construct a deeper network. The normal cells would be repeated $K$ times between the reduction cells where we use $K=1$ for light-weighted network and $K=2$ for the medium one.
      \setlength{\belowcaptionskip}{-5pt}
      }
\label{fig:stacked}
\vspace{0pt}
\end{figure}

\begin{figure*}[t]
\begin{center}
   \includegraphics[width=1\linewidth]{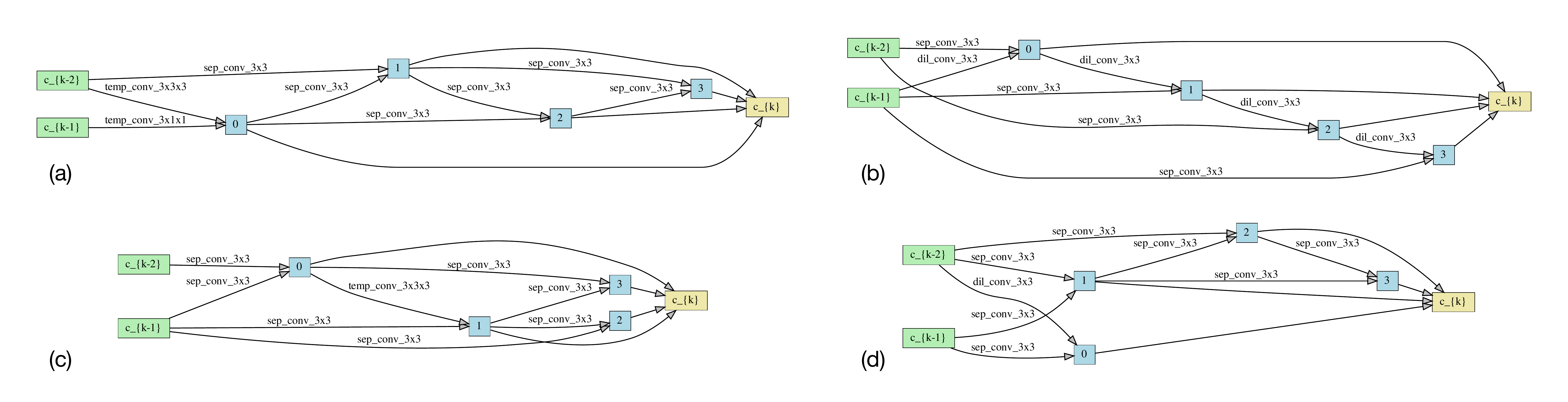}
\end{center}
   \caption{The best architecture we searched on Kinetics-400. (a) The temporal normal cell. (b) The spatial normal cell. (c) The temporal reduction cell. (d) The normal reduction cell. Since the Kinetics-400 dataset is more sensitive to the spatial information in the early layers, the spatial normal cell learns a major path with dilated convolutions which have a large receptive field. In the later layers, since the spatial receptive field is large enough to track the continuous action of an object, the model tends to learn more separate convolutions which have more parameters.
   \setlength{\belowcaptionskip}{-5pt}
   }
\label{fig:searched_arch}
\vspace{0pt}
\end{figure*}

\section{Experiments}
\subsection{Datasets}
We conduct experiments on three video recognition datasets:

\noindent{\textbf{Kinetics-400}}\cite{kay2017kinetics} is a large video recognition dataset focused on human actions with 225,946 training and 18,584 validation videos. The dataset has 400 action classes and each action has around 250-1000 trimmed video clips in the training set. Each video clip lasts around 10 seconds from different YouTube videos. Since the dataset is large enough to train the models from scratch, it has been an important benchmark to evaluate a video recognition method. We search and evaluate our architectures directly on this dataset, achieve state-of-the-art performance while training from scratch.

\noindent{\textbf{Something-Something-V2}}\cite{goyal2017something} is a large collection of densely-labeled video clips that show humans performing pre-defined basic actions with everyday objects. That is a challenging dataset as the activities cannot be inferred merely from individual frames. We retrain our searched architecture from scratch on this dataset to illustrate the generalization of our searched architecture.

\noindent{\textbf{UCF-101}}\cite{ucf101} dataset contains 101 actions with 100+ clips for each action, all the  are taken from only 2.5k distinct videos. Since it is relatively small and requires less computational resources to train, we conduct ablation studies on this dataset.

\begin{table}[t]
	\vspace{0pt}
	\centering
	\small
	\begin{tabular}{l|c|c|c|c|c|c}
		\multicolumn{1}{c|}{Model} & Flow &Frames &\multicolumn{1}{c|}{Pretrain} &   Top-1  & Top-5 & GFLOPs\x views  \\
		\shline
		\demph{I3D \cite{carreira2017quo}} && \demph{32} & \demph{ImageNet} & \demph{72.1} &  \demph{90.3}  & \demph{108 \x} \demph{N/A}  \\
		\demph{Two-Stream I3D \cite{carreira2017quo}}&  \demph{\checkmark}  & \demph{16+16} & \demph{ImageNet} &   \demph{75.7} & \demph{92.0} &   \demph{216~\x} \demph{N/A}  \\
		\demph{S3D-G \cite{Xie_2018_ECCV}}  & \demph{\checkmark} & - & \demph{ImageNet} &   \demph{77.2} & \demph{93.0} &  \demph{143~\x}  \demph{N/A}  \\
		\demph{Nonlocal R50 \cite{wang2017non}}  &  & \demph{32} & \demph{ImageNet} &   \demph{76.5} & \demph{92.6} & \demph{282~\x}  \demph{30}  \\
		\demph{Nonlocal R101 \cite{wang2017non}} &  & \demph{32} & \demph{ImageNet} &   \demph{77.7} & \demph{93.3} & \demph{359~\x}  \demph{30}  \\
		\demph{TSM \cite{lin2019tsm}} &  & \demph{8} & \demph{ImageNet} &   \demph{74.1} & \demph{91.2} &  \demph{65~\x}  \demph{30}  \\
		\hline
		ECO \cite{zolfaghari2018eco}&   & - & - &  70.0 &  89.4 & N/A~\x~N/A  \\
		
		I3D \cite{carreira2017quo}&  \checkmark & 16+16 & - &    71.6 & 90.0 & 216~\x~N/A  \\
		
		R(2+1)D \cite{tran2018closer}&   & 16 & - &   72.0 &  90.0 & 152~\x~115  \\
		R(2+1)D \cite{tran2018closer} & \checkmark & 16+16 & -&  73.9 &  90.9& 304~\x~115  \\
		ir-CSN-152\cite{Tran_2019_ICCV} &  & 32 & - & 76.8  & 92.5 & 96.7~\x~30 \\
		EvaNet \cite{Piergiovanni_2019_ICCV}&  & 8 & - & 77.2  & N/A & N/A\x~30  \\
		ip-CSN-152\cite{Tran_2019_ICCV} &  & 32 & - & 77.8  & 92.8 & 108.8~\x~30 \\
		SlowFast 8\x 8\cite{Feichtenhofer_2019_ICCV}&  & 8+64& - & 77.9  & 93.2 & 106~\x~30  \\
		SlowFast 16\x 8\cite{Feichtenhofer_2019_ICCV} &  & 16+64 & - &79.8 & 93.9 & 234~\x~30  \\
		\hline
		PV-NAS-L & & \textbf{8} & - & 78.7 & 93.5  & 22.14~\x~30 \\
		PV-NAS-M & & \textbf{8} & - & \textbf{81.4} & \textbf{94.2}  & 82.11~\x~30 
	\end{tabular}
	\caption{Comparison with the state-of-the-art on Kinetics-400. Our PV-NAS models achieve state-of-the-art performance. Our PV-NAS-L outperforms SlowFast 1/5 computation cost. Our PV-NAS-M achieves better Top 1 accuracy than methods with least computation cost. Notably, our models use 8-frame clips as input which is clearly a inferiority compared with models trained with more frames eg(SlowFast, TVM). We also report the inference cost with a single clip$\times$the numbers of such views used.
	\setlength{\belowcaptionskip}{-5pt}
    }
	\label{tab:sota:k400}
	\vspace{-10pt}
\end{table}

\subsection{Implementation Details}
\subsubsection{Searching on Kinetics-400}
\noindent{\textbf{Stacked Search Space}}:
As shown in Fig.\ref{fig:stacked}(a), we only stack $L=4$ cells during the search stage, with the order S-NC$\rightarrow$S-RC$\rightarrow$T-NC$\rightarrow$T-RC. Each cell has $6$ nodes and we set the initial channel num as 64.

\noindent{\textbf{Data Augmentation}}:
We directly conduct architecture search on the Kinetics-400 training split. For the temporal dimension, we randomly sample a clip lasts 64 frames from the original 30fps video, then sample 8 frames uniformly in this clip as input. For the spatial dimension, we use a scale jittering range of [128, 164] and randomly crop 112$\times$112 pixels from the frame.

\noindent{\textbf{Hyper Paramaters}}:
We use 32 GPUs for searching and the mini-batch size is 32 clips per GPU. The architecture search finished in 80 epochs, with total time around 20 GPU days. The inner optimization for architecture parameter $a$ used Adam optimizer with learning rate 0.001 and $\beta=(0.0, 0.999)$. The outer optimization for network weights $w$ is implemented by SGD optimizer with momentum 0.9 and weight decay 0.0001. The initial learning rate for the outer optimizer is 0.1 per GPU and it would decay by 10 times for every 20 epochs.

\subsubsection{Discrete Architecture Evaluation}
\noindent{\textbf{Stacked Architecture}}:
We stacked the final searched cells as shown in Fig.\ref{fig:stacked}(b).
We set $N=1$, initial channel as 64 for our light-weight network PV-NAS-L and set $N=2$, initial channel as 100 for the medium network PV-NAS-M. The final searched discrete architecture is shown as Fig.\ref{fig:searched_arch}.

\noindent{\textbf{Data Augmentation}}:
For the Kinetics-400 and Someting-Something-V2 dataset, we use the same temporal sample strategy as the searching stage. We then use a scale jittering range of [256, 320] and randomly crop 224$\times$224 pixels from the frame for the spatial dimension. For the UCF-101 dataset, we use the range [128, 256] for the scale jittering while remain the other settings unchanged.
\noindent{\textbf{Hyper Parameters}}:
We use 32 GPUs to train the network with the mini-batch size 8 clips per GPU. The batch normalization statics individually in each GPU.
For Kinetics-400 and Someting-Something-V2 dataset, we train the network with SGD optimizer for 196 epochs from scratch with momentum 0.9, weight decay 0.0001 and the base learning rate 0.01 per GPU.
We adopt cosine scheduler for learning rate decaying while the first 36 epochs are used for warm up. The dropout probability is set to be 0.4 for the final fully-connected layer.
For UCF-101 dataset, the optimization lasts 128 epochs with 16 warm-up epochs at first. We also use the drop-path regularization with the drop probability 0.1 to prevent the overfitting.
\subsection{Results}
\subsubsection{Results on Kinetics-400}
The model searched on Kinetics-400 has a great fitness on that dataset. Tab.\ref{tab:sota:k400} shows the comparison betwee our PV-NAS model and state-of-the-art results. The lighting model PV-NAS-L achieves a comparable results while significantly reduce the parameters and FLOPs. Compare to TSM\cite{lin2019tsm} which requires a similar computational cost, our model provides 4.6\% higher top-1 accuracy with 8-frame clips as input. The model also achieves 4.0\% better accuracy than the 16-frame TSM model with only 8-frame clips as input, which reduces the computational costs by half. 

The medium model PV-NAS-M also outperforms the previous state-of-the-art methods with lower parameters and FLOPs. Compared to SlowFast\cite{Feichtenhofer_2019_ICCV} with 16$\times$8 frames as input, we could achieve a 1.6\% improvement on the top-1 accuracy. Note that our method shared the same meta-architecture as the SlowPath of the SlowFast network, it illustrates that our machine-optimized method could capture high-quality features include both spatial and temporal information without the auxiliary of FastPath than the manually designed network.

It is also worth to find that our model is significantly better than the evolutionary searched architecture EvaNet\cite{Piergiovanni_2019_ICCV}, which is not consistent with empirical practice on 2D architecture search tasks. There might be two reasons:
1) We have two more categories of cells than the conventional 2D differantiable approaches, that expands the search space quadratically and reduces the gap between the size of the two search spaces.
2) The search space of EvaNet only contains temporal operations but the trimmed videos in Kinetics-400 dataset are more sensitive to the spatial information in the eraly layers. As shown in Fig.\ref{fig:searched_arch}(b), our searched spatial normal cell has a long ``major'' path with dilated convolutions, which provides a large spatial receptive field before aggregate the temporal information.

\subsubsection{Results on Something-Something-V2}
To prove that our searched model has a good generalization, we also conduct experiments on the Something-Something-V2 dataset, which relies more on long-term information compared with the Kinetics dataset. As shown in Tab.\ref{tab:sota:sthsthv2}, while using 8-frame RGB clips as inputs, our model could achieve a better performance than other methods without extra training data.

\begin{table}[t]
	\vspace{0pt}
	\centering
	\small
	\begin{tabular}{l|c|c|c|c}
		\multicolumn{1}{c|}{Model}& Frame &\multicolumn{1}{c|}{Pretrain} & Top-1  & Top-5       \\ 
	    \shline
		
		\hline
		TSN \cite{simonyan2014two} & 16 & Kinetics & 30.0 & 60.5 \\
		TRN Multiscale\cite{zhou2017temporal}  & 8 & ImageNet & 50.9 & 79.3 \\
		TRN Two-Stream\cite{zhou2017temporal} & 8+8 & ImageNet & 55.5 & 83.1 \\
		TSM\cite{lin2019tsm}& 8 & ImageNet & 59.1 & 85.6 \\
		STM\cite{Jiang_2019_ICCV}& 8 & ImageNet & 62.3  & 88.8 \\
		\hline
		PV-NAS-L & 8 & - & {\textbf{62.5}} & {88.4}
	\end{tabular}
	\caption{Comparison with state-of-the-art methods on Sth-Sth-V2. Accuracy is measured on the validation set. The proposed PV-NAS-L model provides a 0.3\% improvement on Top-1 accuracy ( without pretraining). We only train the model with 8-frame RGB clips as input since our model is searched with the same setting.
	\setlength{\belowcaptionskip}{-5pt}
	}
	
	\label{tab:sota:sthsthv2}
\end{table}

\begin{table}[ht]
	\vspace{0pt}
	\centering
	\small
	\begin{tabular}{l|c|c|c|c}
		\multicolumn{1}{c|}{Model} & Epoch& Channel & Top-1  & Top-5 \\ 
		\shline
		
		Baseline(PV-NAS-L)& 80 & 64 & 66.43 & 85.70   \\
		\hline
		Stop at 20 Epoch & 20 & 64 & 64.11 & 84.78 \\
		Stop at 40 Epoch & 40 & 64 & 65.41 & 85.64 \\
		Stop at 60 Epoch & 60 & 64 & 65.80 & 85.69 \\
		\hline
		Half Channel & 80 & 32 & 65.12 & 85.35 \\
		Double Channel & 80 & 128 & 68.51 & 87.1 \\
		\hline
		Random & - & 64& 61.81 & 82.97 \\
	\end{tabular}
	\caption{Ablation studies on UCF-101 dataset. We only report the results trained from scratch for fair comparison.
	\setlength{\belowcaptionskip}{-5pt}
	}
		\label{tab:ablation}
\end{table}

\subsection{Ablation Study}
We conduct ablation study on UCF-101 dataset. Notably, the performance results should not used to directly compare with other works since we do not use extra data for pre-training.

\vspace{0.1cm}
\noindent \textbf{Stop at Different Epochs.}
As mentioned in Sec.\ref{secsec:slr}, we use a scheduled learning rate decay strategy to prevent the architecture from saturating to local minimum. In our implementation, we decay the learning rate for every 20 epochs and the optimization is finished in 80 epochs. To illustrate the effectiveness of the strategy, we rebuild the searched architectures from the 20, 40 and 60 epochs.
The results are shown in Tab.\ref{tab:ablation}, which fits our conclusion well. The model stopped at later stage would have a better performance. 

\vspace{0.1cm}
\noindent \textbf{Different Initial Channels.}
Once the architecture of the searched four cell categories is determined, one option to contribute a bigger network with a higher accuracy is to expand the number of its initial channels. We conduct two experiments which half and double the initial channels of the PV-NAS-L network respectively, shows that a large number of initial channel could always bring a better performance.

\vspace{0.1cm}
\noindent \textbf{Random Wired Network.}
To show that the performance improvements come from the architecture optimization more than the well-defined search space, we random generate 5 network instances from our search space. To forbid the random generated network has little parameters which would lead a bad performance, we ensure that each cell has at least four convolution operations during the random generation. We report the averaged top-k accuracy on the five random generated network, which shows that our searched architecture always has a better performance.

\vspace{-0.2cm}
\section{Conclusion }
In this paper, we propose a innovative and practical way to learn spatio-temporal features for video recognition based on NAS. We first design a comprehensive and effective search space in both space and time, and introduce an efficient gradient-based architecture search algorithm. In order to obtain a set of stable and diverse model architectures, we further propose a new learning rate scheduler. The intensive empirical evaluations show that the propose PV-NAS outperforms all previous hand-crafted networks for video recognition. To our knowledge, it is the first time that the automated designed networks outperform manually designed ones. We also show that the searched network that is crafted for one dataset (i.e,~Kinetics-400) is capable of transferring to a different dataset without performance drop.

\newpage
{\small
\bibliographystyle{splncs04}
\bibliography{egbib}
}
\end{document}